\documentclass[11pt,a4paper]{article}
\usepackage[hyperref]{acl2018}

\usepackage{times}
\usepackage{latexsym}
\usepackage{amssymb}
\usepackage{amsmath}
\usepackage{graphicx}

\usepackage{booktabs}
\usepackage{enumitem}
\setlist[enumerate]{leftmargin=*}
\usepackage{url}
\usepackage{color}

\aclfinalcopy 
\def\aclpaperid{211} 

\title{Dynamic and Static Topic Model\\for Analyzing Time-Series Document Collections}

\author{Rem Hida${}^\dagger$ \quad Naoya Takeishi${}^\ddagger{}^\dagger$ \quad Takehisa Yairi${}^\dagger$ \quad Koichi Hori${}^\dagger$ \\
${}^\dagger$Department of Aeronautics and Astronautics, The University of Tokyo\\
{\tt \{hida,yairi,hori\}@ailab.t.u-tokyo.ac.jp}\\
${}^\ddagger$RIKEN Center for Advanced Intelligence Project, Tokyo, Japan\\
{\tt naoya.takeishi@riken.jp}}

\date{}

\begin{document}

\maketitle

\begin{abstract}
For extracting meaningful topics from texts, their structures should be considered properly. In this paper, we aim to analyze structured time-series documents such as a collection of news articles and a series of scientific papers, wherein topics evolve along time depending on multiple topics in the past, and are also related to each other at each time.
To this end, we propose a \emph{dynamic and static topic model}, which simultaneously considers the dynamic structures of the temporal topic evolution and the static structures of the topic hierarchy at each time. We show the results of experiments on collections of scientific papers, in which the proposed method outperformed conventional models. Moreover, we show an example of extracted topic structures, which we found helpful for analyzing research activities.
\end{abstract}


\section{Introduction}

Probabilistic topic models such as latent Dirichlet allocation (LDA) \citep{blei2003latent} have been utilized for analyzing a wide variety of datasets such as document collections, images, and genes. Although vanilla LDA has been favored partly due to its simplicity, one of its limitations is that the output is not necessarily very understandable because the priors on the topics are independent.
Consequently, there has been a lot of research aimed at improving probabilistic topic models by utilizing the inherent \emph{structures} of datasets in their modeling (see, e.g., \citet{Blei2006DTM,Li2006PAM}; see Section~\ref{related} for other models).

In this work, we aimed to leverage the dynamic and static structures of topics for improving the modeling capability and the understandability of topic models. These two types of structures, which we instantiate below, are essential in many types of datasets, and in fact, each of them has been considered separately in several previous studies. In this paper, we propose a topic model that is aware of both of these structures, namely \emph{dynamic and static topic model} (DSTM).

The underlying motivation of DSTM is twofold.
First, a collection of documents often has \emph{dynamic structures}; i.e., topics evolve along time influencing each other. For example, topics in papers are related to topics in past papers. We may want to extract such dynamic structures of topics from collections of scientific papers for summarizing research activities.
Second, there are also \emph{static structures} of topics such as correlation and hierarchy. For instance, in a collection of news articles, the ``sports'' topic must have the ``baseball'' topic and the ``football'' topic as its subtopic. This kind of static structure of topics helps us understand the relationship among them.

The remainder of this paper is organized as follows. In Section~\ref{related}, we briefly review related work. In Section~\ref{main}, the generative model and the inference/learning procedures of DSTM are presented. In Section~\ref{exp}, the results of the experiments are shown. This paper is concluded in Section~\ref{concl}.


\section{Related Work}\label{related}

Researchers have proposed several variants of topic models that consider the dynamic or static structure.
Approaches focusing on the dynamic structure include dynamic topic model (DTM) \citep{Blei2006DTM}, topic over time (TOT) \citep{Wang2006TOT}, multiscale dynamic topic model (MDTM) \citep{iwata2010online}, dependent Dirichlet processes mixture model (D-DPMM) \citep{lin2010construction}, and infinite dynamic topic model (iDTM) \citep{ahmed2010timeline}.
These methods have been successfully applied to a temporal collection of documents, but none of them take temporal dependencies between multiple topics into account; i.e., in these models, only a single topic contributes to a topic in the future.

For the static structure, several models including correlated topic model (CTM) \citep{lafferty2006correlated}, pachinko allocation model (PAM) \citep{Li2006PAM}, and segmented topic model (STM) \citep{Du2010} have been proposed.
CTM models the correlation between topics using the normal distribution as the prior, PAM introduces the hierarchical structure to topics, and STM uses paragraphs or sentences as the hierarchical structure.
These models can consider the static structure such as correlation and hierarchy between topics. However, most of them lack the dynamic structure in their model; i.e., they do not premise temporal collections of documents.

One of the existing methods that is most related to the proposed model is the hierarchical topic evolution model (HTEM) \citep{song2016discovering}. HTEM captures the relation between evolving topics using a nested distance-dependent Chinese restaurant process.
It has been successfully applied to a temporal collection of documents for extracting structure but does not take multiple topics dependencies into account either.

In this work, we built a new model to overcome the limitation of the existing models, i.e., to examine both the dynamic and static structures simultaneously. We expect that the proposed model can be applied to various applications such as topic trend analysis and text summarization.

\begin{table}[t]
  \centering
  {\fontsize{9.25pt}{9.25pt}\selectfont\begin{tabular}{lp{6cm}}
    \toprule
    $D^{t}$ & number of documents at epoch $t$\\
    $n_d^{t}$ & number of words in the $d$-th doc. at epoch $t$\\
    $w^{t}_{d,i}$ & the $i$-th word in the $d$-th doc. at epoch $t$\\
    $K$ & total number of subtopics\\
    $S$ & number of supertopics\\
    $y_{d,i}^{t}$ & supertopic of $w^{t}_{d,i}$\\
    $z_{d,i}^{t}$ & subtopic of $w^{t}_{d,i}$\\
    ${}^{1}\theta_{d}^{t}$ & multinomial distribution over supertopics for the $d$-th doc. at epoch $t$\\
    ${}^{2}\theta_{d,s}^{t}$ & multinomial distribution over subtopics for the $d$-th doc. in $s$-th supertopic at epoch $t$\\
    $\phi_{k}^{t}$ & multinomial distribution over words for the $k$-th subtopic at epoch $t$ \\
    ${}^{2}\alpha_{s}^{t}$ & static structure weight (prior of ${}^{2}\theta_{d,s}^t$) \\
    $\beta^{t}$ & dynamic structure weight between topics at time $t-1$ and those at epoch $t$\\
    \bottomrule
  \end{tabular}}%
  \vspace{-0.5em}
  \caption{Notations in the proposed model.}
  \label{notation}
\end{table}


\section{Dynamic and Static Topic Model}\label{main}

In this section, we state the generative model of the proposed method, DSTM. Afterward, the procedure for inference and learning is presented. Our notations are summarized in Table~\ref{notation}.


\subsection{Generative Model}

In the proposed model, DSTM, the dynamic and static structures are modeled as follows.

\begin{figure}[t]
  \centering
  \includegraphics[width=6.4cm]{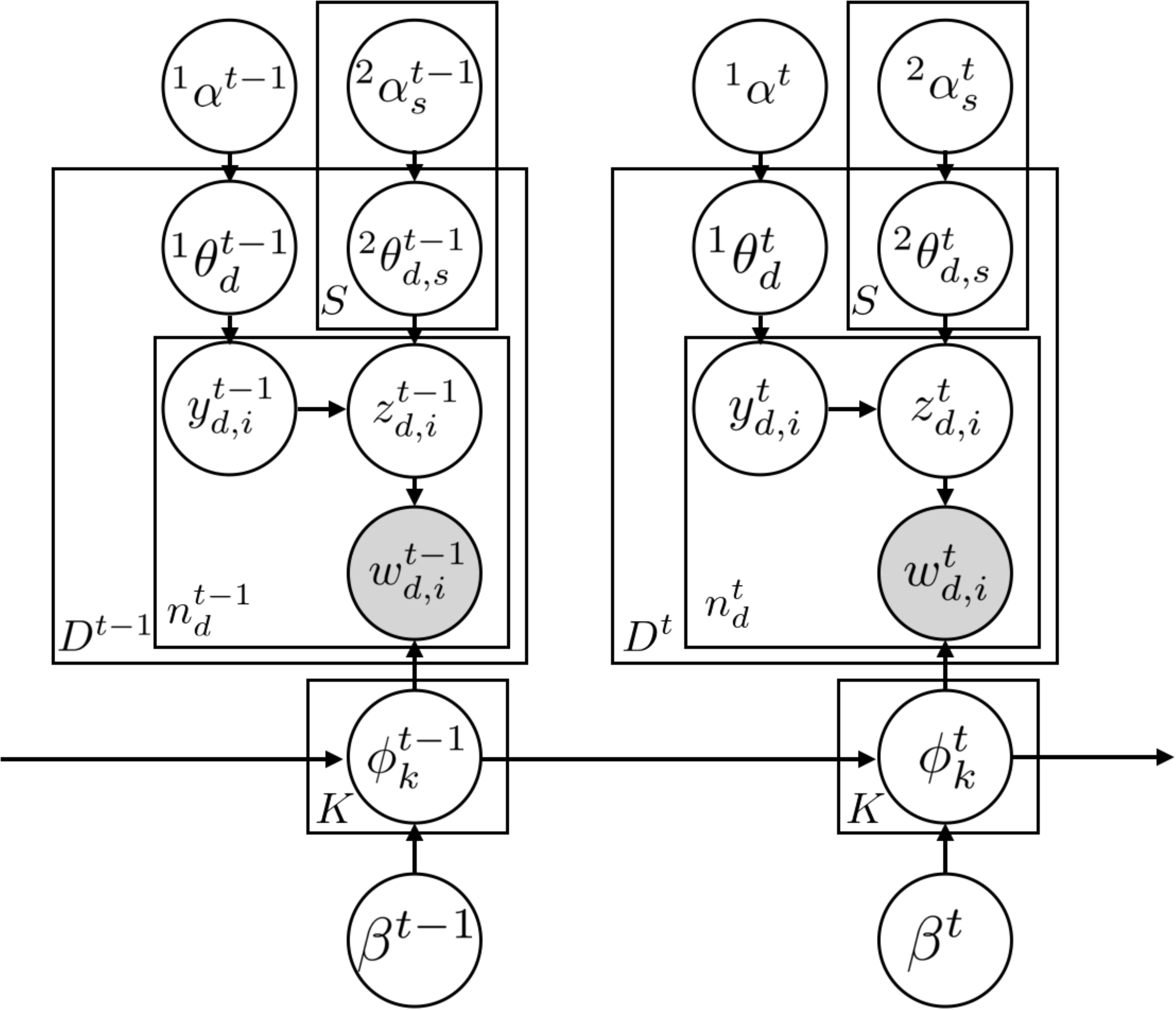}
  \vspace{-1em}
  \caption{Graphical model of the proposed model for epochs $t-1$ and $t$.}
  \label{graphical_model}
\end{figure}

\paragraph{Dynamic Structure}
We model the temporal evolution of topic-word distribution by making it proportional to a weighted sum of topic-word distributions at the previous time (epoch), i.e.,
\begin{equation}
  \phi^{t}_{k} \sim \mathrm{Dirichlet} \left( \sum_{k'=1}^K \beta^{t}_{k,k'} {\phi}^{t-1}_{k'} \right),
  \label{trans}
\end{equation}
where $\phi^t_k$ denotes the word distribution of the $k$-th topic at the $t$-th time-epoch, and $\beta^t_{k,k'}$ is a weight that determines the dependency between the $k$-th topic at epoch $t$ and the $k'$-th topic at epoch $t-1$.

\paragraph{Static Structure}
We model the static structure as a hierarchy of topics at each epoch. We utilize the supertopic-subtopic structure as in PAM \citep{Li2006PAM}, where the priors of topics (subtopics) are determined by their supertopic.

\paragraph{Generative Process}
In summary, the generative process at epoch $t$ is as follows.
\begin{enumerate}[topsep=1pt]
  \setlength{\parskip}{0cm} 
  \setlength{\itemsep}{0cm} 
  \item For each subtopic $k=1,..,K$ ,
  \begin{enumerate}
    \item Draw a topic-word distribution \\
    $\phi_{k}^{t} \sim \mathrm{Dirichlet}(\sum_{k'}\beta^t_{k,k'}{\phi}^{t-1}_{k'})$.
  \end{enumerate}
  \item For each document $d=1,...,D^{t}$,
  \begin{enumerate}
    \item Draw a supertopic distribution \\
    ${}^{1}\theta_{d}^{t}\sim \mathrm{Dirichlet}({}^{1}\alpha^{t})$.
    \item For each supertopic $s=1,...,S$,
    \begin{enumerate}
      \item Draw  a subtopic distribution \\
      ${}^{2}\theta_{d,s}^{t}\sim \mathrm{Dirichlet}({}^{2}\alpha_{s}^{t})$.
    \end{enumerate}
    \item For each word $i=1,...,n_d^{t}$,
    \begin{enumerate}
      \item  Draw a supertopic-word assignment \\
      $y_{d,i}^{t}\sim \mathrm{Multinomial}({}^{1}\theta_{d}^{t})$.
      \item Draw a subtopic-word assignment \\
      $z_{d,i}^{t}\sim \mathrm{Multinomial}({}^{2}\theta^{t}_{d,y^{t}_{d,i}})$.
      \item Draw a word-observation \\
      $w_{d,i}^{t}\sim \mathrm{Multinomial}(\phi_{{z}_{d,i}^{t}}^{t})$.
    \end{enumerate}
  \end{enumerate}
\end{enumerate}
Note that the above process should be repeated for every epoch $t$. The corresponding graphical model is presented in Figure~\ref{graphical_model}.


\subsection{Inference and Learning}

Since analytical inference for DSTM is intractable, we resort to a stochastic EM algorithm \citep{Andrieu2003} with the collapsed Gibbs sampling \citep{Griffiths5228}. However, such a strategy is still much costly due to the temporal dependencies of $\phi$. Therefore, we introduce a further approximation; we surrogate $\phi_{k'}^{t-1}$ in Eq.~\eqref{trans} by its expectation $\hat{\phi}_{k'}^{t-1}=\mathbb{E}[\phi_{k'}^{t-1}]$. This compromise enables us to run the EM algorithm \emph{for each} epoch in sequence from $t=1$ to $t=T$ without any backward inference.
\textcolor{black}{In fact, such approximation technique is also utilized in the inference of MDTM \citep{iwata2010online}.}

\textcolor{black}{Note that the proposed model has a moderate number of hyperparameters to be set manually, and that they can be tuned according to the existing know-how of topic modeling. This feature makes the proposed model appealing in terms of inference and learning.}

\paragraph{E-step}
In E-step, the supertopic/subtopic assignments are sampled. Given the current state of all variables except ${y}_{d,i}^{t}$ and ${z}_{d,i}^{t}$, new values for them should be sampled according to
\begingroup\makeatletter\def\f@size{9.5}\check@mathfonts
\begin{equation}\begin{aligned}
  &p({y}_{d,i}^{t}=s,{z}_{d,i}^{t}=k \mid w^t,y^{t},z^{t},\Phi^{t-1},{}^{1}\alpha^{t},{}^{2}\alpha^{t},\beta^{t})\\
  &\quad\propto \frac{n^{t}_{d,s\backslash i}+{}^{1}\alpha_{s}^{t}}{n^{t}_{d\backslash i}+\sum_{s=1}^{S}{}^{1}\alpha_{s}^{t}}
  \cdot
  \frac{n^{t}_{d,s,k\backslash i}+{}^{2}\alpha_{s,k}^{t}}{n^{t}_{d,s\backslash i}+\sum_{k=1}^{K}{}^{2}\alpha_{s,k}^{t}}
  \\
  &\qquad\cdot \frac{n^{t}_{k,v\backslash i}+\sum_{k'=1}^{K}\beta_{k,k'}^{t} \hat{\phi}^{t-1}_{k',v}}{n^{t}_{k\backslash i}+\sum_{k'=1}^{K}\beta_{k,k'}^{t}},
\end{aligned}\end{equation}
\endgroup
where $n_{k,v}^{t}$ denotes the number of tokens assigned to topic $k$ for word $v$ at epoch $t$, $n_{k}^{t}\mathalpha{=}\sum_{v}n_{k,v}^{t}$, and $n_{d,s}^{t}$ and $n_{d,s,k}^{t}$ denote the number of tokens in document $d$ assigned to supertopic $s$ and subtopic $k$ (via $s$), at epoch $t$ respectively. Moreover, $n_{\cdot \backslash i}^{t}$ denotes the count yielded excluding the $i$-th token.

\paragraph{M-step}
In M-step, ${}^{2}\alpha^{t}$ and $\beta^{t}$ are updated using the fixed-point iteration \citep{minka2000estimating}.
\begingroup\makeatletter\def\f@size{8.5}\check@mathfonts
\begin{align}
  ({}^{2}\alpha_{s,k}^{t})^* &=
  {}^{2}\alpha_{s,k}^{t}
  \frac{\sum_{d=1}^{D^{t}} \Psi(n_{d,s,k}^{t} + {}^{2}\alpha_{s,k}^{t}) - \Psi({}^{2}\alpha_{s,k}^{t})}
  {\sum_{d=1}^{D^{t}} \Psi(n_{d,s}^{t}+{}^{2}\alpha_{s}^{t}) - \Psi({}^{2}\alpha_{s}^{t})},\\
  (\beta^{t}_{k,k'})^* &=
  \beta^{t}_{k,k'}
  \frac{\sum_{v}\hat{\phi}^{t-1}_{k', v} B^{t}_{k', v}}
  {\Psi(n^{t}_{k} + \sum_{k'}\beta^{t}_{k,k'}) - \Psi(\sum_{k'}\beta^{t}_{k,k'})}.
\end{align}
\endgroup
Here, $\Psi$ is the digamma function, ${}^{2}\alpha_{s}^{t} \mathalpha{=} \sum_{k}{}^{2}\alpha_{s,k}^{t}$, and
\begingroup\makeatletter\def\f@size{9.2}\check@mathfonts
\begin{equation*}
  B^{t}_{k', v} = \Psi \Bigl( n^{t}_{k, v}+\sum_{k'}\beta^{t}_{k,k'}\hat{\phi}^{t-1}_{k', v} \Bigr) -\Psi \Bigl( \sum_{k'}\beta^{t}_{k,k'}\hat{\phi}^{t-1}_{k', v} \Bigr).
\end{equation*}
\endgroup

\paragraph{Overall Procedure}
The EM algorithm is run for each epoch in sequence; at epoch $t$, after running the EM until convergence, $\hat{\phi}^{t}_{k,v}$ is computed by
\begin{equation*}
  \hat{\phi}^{t}_{k,v}=\frac{n^{t}_{k,v}+\sum_{k'}\beta^{t}_{k,k'}\hat{\phi}^{t-1}_{k',v}}{n^{t}_{k}+\sum_{k'}\beta^{t}_{k,k'}},
\end{equation*}
and then this value is used for the EM at the next epoch $t+1$. Moreover, see Supplementary~\ref{para_est} for the computation of the statistics of the other variables.


\section{Experiments}\label{exp}


\subsection{Datasets}

We used two datasets comprising technical papers: \textbf{NIPS} \citep{dataset} and \textbf{Drone} \citep{dataset_drone}. \textbf{NIPS} is a collection of the papers that appeared in NIPS conferences. \textbf{Drone} is a collection of abstracts of papers on unmanned aerial vehicles (UAVs) and was collected from related conferences and journals for surveying recent developments in UAVs. The characteristics of those datasets are summarized in Table~\ref{nips_drones}. See Supplementary~\ref{prepro} for the details of data preprocessing.

\begin{table}[t]
  \centering
  \begin{tabular}{ccc}
  \toprule
    & \textbf{NIPS} & \textbf{Drone}\\
  \midrule
    Date&1987--1999&2009--2016\\
    \# Documents&1,740&1,035\\
    \# Vocabulary&11,443&3,442\\
    \# Tokens&2,271,087&68,305\\
  \bottomrule
  \end{tabular}
  \vspace{-0.5em}
  \caption{Summary of the datasets.}
  \label{nips_drones}
\end{table}

\begin{table*}[t]
  \centering
  {\small\scalebox{0.92}[1.0]{\begin{tabular}{ccccccccc}
    \toprule
      &&&& \textbf{NIPS} &&& \textbf{Drone} &\\
    \cmidrule(lr){4-6} \cmidrule(lr){7-9}
      &static&dynamic&K30 (S15)&K40 (S20)&K50 (S25)&K15 (S3)&K20 (S3)&K25 (S3)\\
    \midrule
      LDA&-&- & 1455.6 (16.7) &1407.3 (15.9)&1374.6 (16.8)&1624.3 (191.1)&1634.8 (189.1)&1644.7 (193.0)\\
      PAM&\checkmark&-&  1455.1 (18.2) &1407.0 (17.5)&1376.9 (16.7)&1587.4 (185.1)&1589.9 (191.4)&1590.8 (186.8)\\
      DRTM&-&\checkmark& 1380.7 (18.5) &1308.6 (17.5)&1253.9 (17.9)&1212.5 (153.2)&1206.1 (148.0)&1201.2 (143.5)\\
      DSTM&\checkmark&\checkmark &\bf{1378.7 (16.5)}&\bf{1301.0 (17.9)}&\bf{1247.3 (17.2)}&\bf{1194.2 (148.2)}&\bf{1180.0 (147.0)}&\bf{1171.6 (141.4)}\\
  \bottomrule
  \end{tabular}}}%
  \vspace{-0.5em}
  \caption{Means (and standard deviations) of PPLs averaged over all epochs for each dataset with different values of $K$ and $S$. The proposed method, DSTM, achieved the smallest PPL.}
  \label{compared ppl}
\end{table*}

\begin{figure*}[t]
\centering
  \includegraphics[width=15cm]{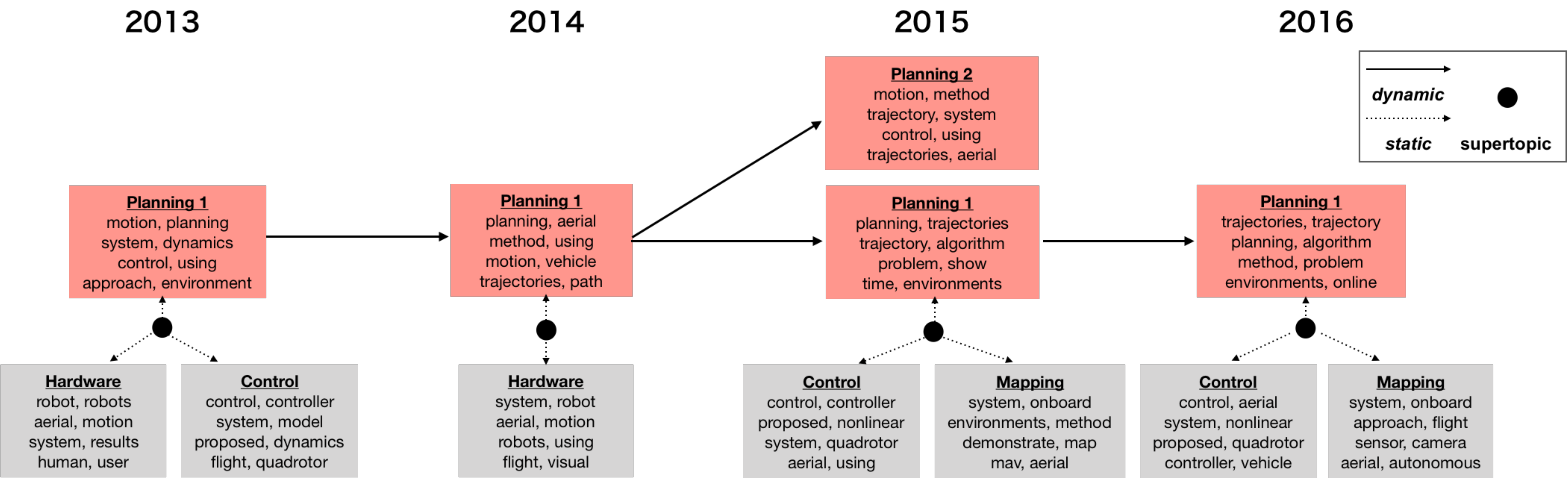}
  \vspace{-1em}
  \caption{Part of the topic structure extracted from \textbf{Drone} dataset using the proposed method. The solid arrows denote the temporal evolution of ``planning'' topics. The dotted arrows mean that ``planning'' topics are related to ``hardware'', ``control'', and ``mapping'' topics via some supertopics (filled circles).}
  \label{topic_trans}
\end{figure*}


\subsection{Evaluation by Perplexity}

First, we evaluate the performance of the proposed method quantitatively using perplexity (PPL):
\begingroup\makeatletter\def\f@size{9.5}\check@mathfonts
\begin{equation*}
  \mathrm{PPL}=\exp\left(-\frac{\sum_{d=1}^{D} \sum_{w_d^\text{test}}\log{p(w_{d,i}}|{\mathcal M})}{\sum_{d=1}^{D}n_{d}^\text{test}}\right).
\end{equation*}
\endgroup
For each epoch, we used 90\% of tokens in each document for training and calculated the PPL using the remaining 10\% of tokens. We randomly created 10 train-test pairs and evaluated the means of the PPLs over those random trials.
We compared the performance of DSTM to three baselines: LDA \citep{blei2003latent}, PAM \citep{Li2006PAM}, and the proposed model without the static structure, which we term DRTM. See Supplementary~\ref{hypara} on their hyperparameter setting.

The means of the PPLs averaged over all epochs for each dataset with different values $K$ are shown in Table~\ref{compared ppl}.
In both datasets with every setting of $K$, the proposed model, DSTM, achieved the smallest PPL, which implies its effectiveness for modeling a collection of technical papers.
\textcolor{black}{For clarity, we conducted paired t-tests between the perplexities of the proposed method and those of the baselines. On the differences between DSTM and DRTM, the p-values were $4.2\times10^{-2}$ ($K=30$), $7.9\times10^{-5}$ ($K=40$), and $6.4\times10^{-7}$ ($K=50$) for the \textbf{NIPS} dataset, and $1.3\times10^{-4}$ ($K=15$), $8.8\times10^{-5}$ ($K=20$), and $4.9\times10^{-6}$ ($K=25$) for the \textbf{Drone} dataset, respectively.}
It is also noteworthy that DRTM shows more significant improvement relative to LDA than PAM does. This suggests that the dynamic structure with multiple-topic dependencies is essential for datasets of this kind.


\subsection{Analysis of Extracted Structure}

We examined the topic structures extracted from the \textbf{Drone} dataset using DSTM.
In Figure~\ref{topic_trans}, we show a part of the extracted structure regarding planning of the UAV's path and/or movement. We identified ``planning'' topics by looking for keywords such as ``trajectory'' and ``motion.'' In Figure~\ref{topic_trans}, each node is labeled with eight most probable keywords. Moreover, solid arrows (dynamic relations) are drawn if the corresponding $\beta^{t}_{k,k'}$ is larger than 200, and dotted arrows (static relations) are drawn between a supertopic and subtopics with the two or three largest values of ${}^{2}\alpha^{t}_{s,k}$.

Looking at the dynamic structure, we may see how research interest regarding planning has changed. For example, word ``online'' first emerges in the ``planning'' topic in 2016. This is possibly due to the increasing interest in real-time planning problems, which is becoming feasible due to the recent development of on-board computers.
In regard to the static structures, for example, the ``planning'' topic is related to the ``hardware'' and ``control'' topics in 2013 and 2014, whereas it is also related to the ``mapping'' topic in 2015 and 2016. Looking at these static structures, we may anticipate how research areas are related to each other in each year. In this case, we can anticipate that planning problems are combined with mapping problems well in recent years.
Note that we cannot obtain these results unless the dynamic and static structures are considered simultaneously.


\section{Conclusion}\label{concl}

In this work, we developed a topic model with dynamic and static structures. We confirmed the superiority of the proposed model to the conventional topic models in terms of perplexity and analyzed the topic structures of a collection of papers.
Possible future directions of research include automatic inference of the number of topics and application to topic trend analysis in various domains.


\bibliographystyle{acl_natbib}
\bibliography{ref}

\clearpage
\appendix{{\LARGE Supplementary}}
\section{Parameter Estimation}\label{para_est}

The means of $\theta^{t}_{d,s}$ and $\theta^{t}_{d,s,k}$ can be obtained by
\begingroup\makeatletter\def\f@size{9.5}\check@mathfonts
\begin{align*}
  {}^{1}\hat{\theta}^{t}_{d,s} &= \frac{n^{t}_{d,s}+{}^{1}\alpha_{s}^{t}}{n^{t}_{d}+\sum_{s=1}^{s=S}{}^{1}\alpha_{s}^{t}}\quad\text{and}\\
  {}^{2}\hat{\theta}^{t}_{d,s,k} &= \frac{n^{t}_{d,s,k}+{}^{2}\alpha_{s,k}^{t}}{\sum_{k=1}^{K}(n^{t}_{d,s,k}+{}^{2}\alpha_{s,k}^{t})}.
\end{align*}
\endgroup

\section{Data Preprocessing}\label{prepro}

We obtained and preprocessed the \textbf{NIPS} and the \textbf{Drone} datasets by the following procedure.
For \textbf{NIPS} dataset, we downloaded its bag-of-words representations, which are available online .\footnote{\url{https://archive.ics.uci.edu/ml/dataset/NIPS+Conference\\+Papers+1987-2015}.} In this dataset, the stop words and the words that appeared less than 50 times in 1987--2015 are originally removed.
For the \textbf{Drone} dataset, we used a spreadsheet\footnote{\url{https://goo.gl/cCoCwL}} that lists UAV-related papers and downloaded abstracts of papers according to the list. In this dataset, the stop words and words that appeared less than four times in 2009--2016 were removed.


\section{Hyperparameter Setting}\label{hypara}

We set the hyperparameters of the proposed model and the baselines as follows.
For LDA, we used the symmetric Dirichlet priors with $\alpha=0.1$ and $\beta = 0.1$.
For PAM, we used symmetric Dirichlet priors with ${}^{1}\alpha = 0.1$ and $\beta = 0.1$ and the initial values of ${}^{2}\alpha = 1.0$.
For DRTM, we used symmetric Dirichlet priors with $\alpha = 0.1$ and initial values of $\beta^{t}_{k,k'} = 100$ if $k = k'$, and $0.1$ otherwise.
For DSTM, we used the symmetric Dirichlet priors with ${}^{1}\alpha = 0.1$ and initial values of ${}^{2}\alpha = 1.0$ and $\beta^{t}_{k,k'} = 100$ if $k = k'$, and $0.1$ otherwise.
Moreover, we randomly initialized the topic assignments and ran the collapsed Gibbs sampler for 500 iterations for every model.

\end{document}